\documentclass{article}
\usepackage{iclr2026_conference,times}

\usepackage{amsmath,amsfonts,bm}

\def\eqref#1{equation~\ref{#1}}

\def\1{\bm{1}}

\DeclareMathAlphabet{\mathsfit}{\encodingdefault}{\sfdefault}{m}{sl}
\SetMathAlphabet{\mathsfit}{bold}{\encodingdefault}{\sfdefault}{bx}{n}

\usepackage{tikz}
\usepackage{collcell}
\usepackage{microtype}
\usepackage{graphicx}
\usepackage{subfigure}
\usepackage{booktabs}
\usepackage{colortbl}
\usepackage{color}
\usepackage{listings}
\usepackage{pdfpages}
\usepackage{minted}
\usepackage{bbding}
\usepackage{xspace}
\usepackage{colortbl,pifont}

\usepackage{subcaption}
\usepackage{hyperref}
\usepackage{wrapfig}
\usepackage{amsmath}
\usepackage{amssymb}
\usepackage{mathtools}
\usepackage{amsthm}
\usepackage{multirow}
\usepackage{enumitem}
\usepackage{dblfloatfix}
\hypersetup{
    colorlinks,
    urlcolor=black,
    linkcolor=black,
    citecolor=black,
}
\usepackage[capitalize,noabbrev]{cleveref}
\usepackage{tabu}
\usepackage{listings}
\usepackage{xcolor} 
\definecolor{commentsColor}{rgb}{0.497, 0.497, 0.497}
\definecolor{keywordsColor}{rgb}{0.0, 0.0, 0.635}
\definecolor{stringColor}{rgb}{0.558, 0.0, 0.135}

\usepackage[utf8]{inputenc}
\usepackage{listings}
\usepackage{xcolor}
\usepackage{tcolorbox}

\lstdefinestyle{mystyle}{
  basicstyle=\ttfamily\small,
  keywordstyle=\color{blue!70!black},
  commentstyle=\color{gray},
  stringstyle=\color{green!50!black},
  showstringspaces=false,
  breaklines=true,
  numbers=none
}

\newcommand{\name}{TGM\xspace}

\theoremstyle{plain}
\newtheorem{theorem}{Theorem}[section]

\theoremstyle{definition}
\newtheorem{definition}[theorem]{Definition}

\theoremstyle{remark}

\renewcommand{\vec}[1]{\ensuremath{\mathbf{#1}}}

\newcommand{\mat}[1]{\ensuremath{\mathbf{#1}}}

\usepackage{pifont}

\definecolor{firstcolour}{HTML}{265D86}
\definecolor{secondcolour}{HTML}{B65A3E}
\definecolor{MutedRed}{HTML}{BF616A}
\definecolor{MonospaceDark}{HTML}{1F232A}

\newcommand{\cfirst}[1]{\textcolor{firstcolour}{\textbf{#1}}}
\newcommand{\csecond}[1]{\textcolor{secondcolour}{\textbf{\textit{#1}}}}
\newcommand{\cna}{{\scriptsize {\textcolor{MonospaceDark}{\ding{53} }}}} 

\newcommand{\hfirst}{\textcolor{firstcolour}{\textbf{first}}}
\newcommand{\hsecond}{\textcolor{secondcolour}{\textbf{\textit{second}}}}

\title{TGM: A Modular and Efficient Library
for Machine Learning on Temporal Graphs}

\author{Jacob Chmura\textsuperscript{1,2}\thanks{Equal contributions, emails: \texttt{\{jacob.chmura, shenyang.huang\} @mail.mcgill.ca} }\quad Shenyang Huang\textsuperscript{3,1,2}\footnotemark[1] \quad Tran Gia Bao Ngo \textsuperscript{4} \quad  Ali Parviz \textsuperscript{1} \\ 
\textbf{Farimah Poursafaei}\textbf{\textsuperscript{1,2}} \quad \textbf{Jure Leskovec}\textbf{\textsuperscript{6,7}} \quad  \textbf{Michael Bronstein}\textbf{\textsuperscript{3,5}} \\
\textbf{Guillaume Rabusseau}\textbf{\textsuperscript{1,8,9}} \quad \textbf{Matthias Fey}\textbf{\textsuperscript{7}} \quad \textbf{Reihaneh Rabbany}\textbf{\textsuperscript{1,2,9}} \\
\textsuperscript{1}Mila - Quebec AI Institute, 
\textsuperscript{2}School of Computer Science, McGill University\\
\textsuperscript{3}University of Oxford,
\textsuperscript{4}Department of computer science, University of Manitoba\\
\textsuperscript{5}AITHYRA,
\textsuperscript{6}Stanford University
\textsuperscript{7}Kumo.AI\\
\textsuperscript{8}DIRO, Université de Montréal,
\textsuperscript{9}CIFAR AI Chair\\
}

\iclrfinalcopy 

\begin{document}

\maketitle
\begin{abstract}

Well-designed open-source software drives progress in Machine Learning~(ML) research. While static graph ML enjoys mature frameworks like PyTorch Geometric and DGL, ML for temporal graphs (TG), networks that evolve over time, lacks comparable infrastructure. Existing TG libraries are often tailored to specific architectures, hindering support for diverse models in this rapidly evolving field. Additionally, the divide between continuous- and discrete-time dynamic graph methods (CTDG and DTDG) limits direct comparisons and idea transfer. To address these gaps, we introduce Temporal Graph Modelling (\name), a research-oriented library for ML on temporal graphs, the first to unify CTDG and DTDG approaches. \name offers first-class support for dynamic node features, time-granularity conversions, and native handling of link-, node-, and graph-level tasks. Empirically, \name achieves an average 7.8× speedup across multiple models, datasets, and tasks compared to the widely used DyGLib, and an average 175× speedup on graph discretization relative to available implementations. Beyond efficiency, we show in our experiments how \name unlocks entirely new research possibilities by enabling dynamic graph property prediction and time-driven training paradigms, opening the door to questions previously impractical to study.

\vspace{0.4em}
\raisebox{-0.2\height}{\hspace{0.7cm}\includegraphics[width=0.44cm]{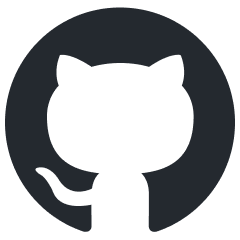}} \small \hspace{0.04cm}\textbf{\mbox{Code:}} \href{https://github.com/tgm-team/TGM}{tgm-team/tgm}
\hspace{5em}
\raisebox{-0.2\height}{\includegraphics[width=0.37cm]{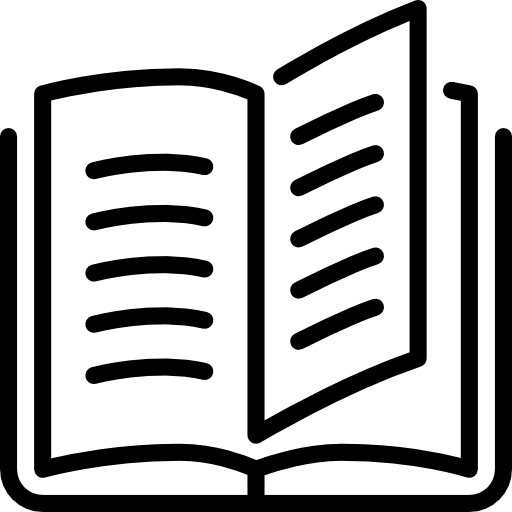}} \small \textbf{\mbox{ Documentation:}} \href{https://tgm.readthedocs.io/en/latest}{tgm.readthedocs.io}

\end{abstract}


\section{Introduction and Motivation}

Advances in machine learning are driven by open, easy-to-use libraries that let researchers focus on developing frontier architectures. For example, deep learning research was propelled by Caffe~\citep{DBLP:conf/mm/JiaSDKLGGD14}, TensorFlow~\citep{DBLP:conf/osdi/AbadiBCCDDDGIIK16} and PyTorch~\citep{DBLP:conf/nips/PaszkeGMLBCKLGA19}. Similarly, developments in graph machine learning~\citep{kipfsemi,velivckovic2017graph, DBLP:journals/corr/abs-2012-09699,rampavsek2022recipe} are accelerated by libraries such as PyG~\citep{DBLP:journals/corr/abs-1903-02428, DBLP:journals/corr/abs-2507-16991} and DGL~\citep{wang2019deep}. However, both PyG and DGL are designed for static graphs and cannot capture the temporal dynamics of networks, known as Temporal Graphs~(TGs). Real-world examples include transaction~\citep{DBLP:conf/nips/ShamsiVKGA22}, social~\citep{huang2023temporal}, trade~\citep{poursafaei2022towards}, and communication networks~\citep{DBLP:journals/corr/abs-2011-13085} among others. 

Recently, Temporal Graph Learning (TGL) has emerged to capture both spatial and temporal dependencies in networks~\citep{cornell2025power,cao2020spectral,han2014chronos}. The field has seen growth with high-impact, cross-domain applications, such as LinkedIn's LiGNN system~\citep{borisyuk2024lignn} for user recommendations and mobility modelling that informed COVID-19 policy decisions~\citep{chang2021mobility}. Unlike static graph ML, TGL must treat time as a first-class signal, making timestamps central to modelling and data processing. Despite research progress, software infrastructure has not kept pace.

\textbf{Limitations of existing libraries.} Current TG libraries~\citep{dygfomer,DBLP:conf/cikm/RozemberczkiSHP21} are narrow in scope: many implement only a single algorithm family~\citep{10.1145/3620665.3640414, DBLP:conf/sc/ZhouZSKP23} and most lack extensibility, resulting in a fragmented ecosystem. For instance, TGL~\citep{DBLP:journals/corr/abs-2203-14883}, DistTGL~\citep{DBLP:conf/sc/ZhouZSKP23} and TGLite~\citep{10.1145/3620665.3640414} are optimized for temporal message passing architectures~\citep{tgn,tgat} but do not support emerging transformer-based approaches~\citep{dygfomer,DBLP:journals/corr/abs-2507-11836}. Also, none provide time conversion operations which are critical for analyzing temporal granularity in TGs~\citep{utg}. Finally, existing libraries fall short on usability features needed to foster reproducible research such as profiling tools, test suites, and modular abstractions (see Table~\ref{tab:features}).

\textbf{Motivation for a unified framework.} Unlike NLP, where the transformer serves as a canonical architecture~\citep{vaswani2017attention}, TGL lacks a standard model family. This leads to fragmented and error-prone experimentation: continuous- and discrete-time models require entirely different data pipelines, while core operations such as temporal neighbor sampling and negative edge construction are implemented inconsistently. Without a unified framework, the community faces difficulties in fair benchmarking, rapid prototyping, and combining ideas across approaches.

\textbf{Our solution.} We introduce \name, a modular and efficient framework for TGL research. \name introduces several firsts: native support for node events, a generic hook mechanism that standardizes TG transformations, and unified support for both continuous- and discrete-time graphs, ending the long-standing separation between the two lines of research~\citep{tgn,DBLP:conf/kdd/YouDL22}. Node events naturally capture phenomena like social media posts or other user activity in real-world networks~\citep{DBLP:journals/jmlr/KazemiGJKSFP20}. These abstractions unify diverse TG pipelines, lowering the barrier for practitioners and accelerating innovation. Beyond flexibility, \name delivers efficiency: $7.8\times$ faster than DyGLib on standard TG models and an average speedup of $175\times$ on graph discretization.


\begin{figure}
    \centering
    \includegraphics[width=0.7\textwidth]{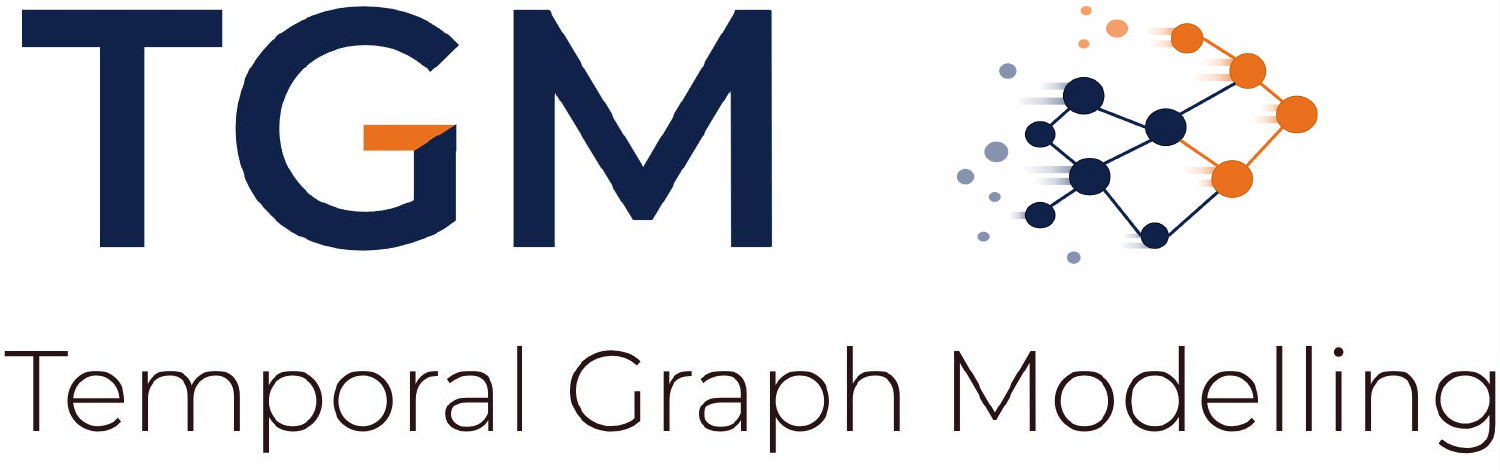}
    \vspace{-1.2em}
\end{figure}

\begin{figure*}[t]
  \centering
\includegraphics[width=0.9\linewidth]{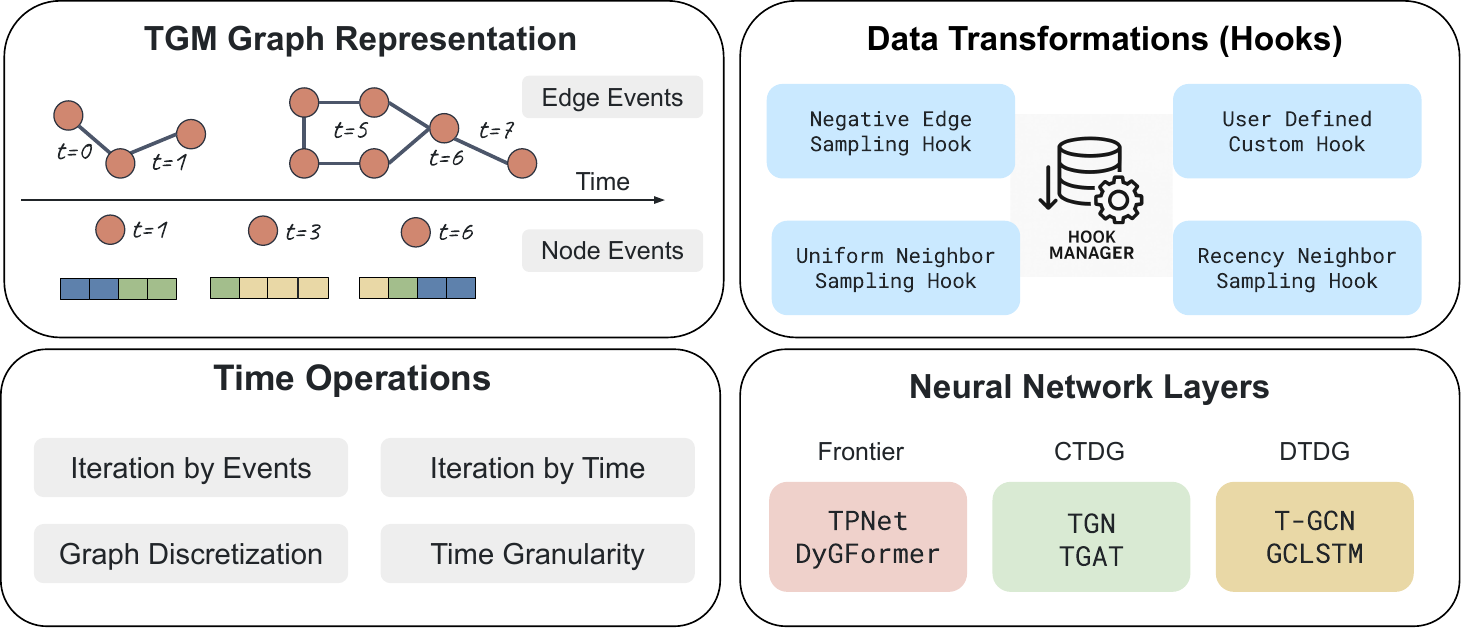}
  \caption{Overview of \name features. \name has native support for node events and unified continuous- and discrete-time graph iteration (left). Generic hooks formalize common TG transformations (top-right). \name supports a broad range of temporal graph learning methods (bottom-right).}
  \label{fig:paradigm}
\end{figure*}

In summary, the key properties of \name are:
\begin{itemize}[leftmargin=*, noitemsep, topsep=0.2pt]
    \item \textbf{First unified library for TG.} \name is the first library to support both continuous- and discrete-time graphs, treating them as distinct views of the same underlying data. We implement 8 methods from both CTDG and DTDG literature, including frontier models.

    \item \textbf{Time as a first-class citizen.} Time operations are central to TGs. \name natively incorporates time granularity into its API, with built-in support for graph discretization and snapshot iteration.

    \item \textbf{Efficiency.} Our experiments show that \name achieves an average $7.8\times$ faster end-to-end training than DyGLib, and $175\times$ faster graph discretization compared with existing implementations.

    \item \textbf{Research-oriented.}  Designed for rapid prototyping, \name emphasizes modularity and ease-of-use. Its novel hook mechanism standardizes temporal graph transformations while supporting the broadest range of TG tasks: link, node, and graph-level prediction. 
\end{itemize}

\section{Related Work}\label{sec:rw}

\textbf{CTDG Methods.}
Continuous-time Dynamic Graph (CTDG) methods process temporal graphs as streams of timestamped edge events. TGAT~\citep{tgat} pioneered inductive representation learning on temporal graphs, and TGN~\citep{tgn} generalized this approach into a widely adopted framework, with TGAT as a special case. Both rely on temporal neighbor sampling for message passing. More recently, DyGLib~\citep{dygfomer} emerged as a popular library, introducing DyGFormer, one of the first transformer-based CTDG architectures inspired by their success in time series, NLP, and vision~\citep{vaswani2017attention,DBLP:conf/naacl/DevlinCLT19,DBLP:conf/iclr/DosovitskiyB0WZ21}. Despite these advances, \citet{edgebank} exposed flaws in prior evaluation and proposed EdgeBank, a strong heuristic baseline for link prediction. To address reproducibility, \citet{huang2023temporal} introduced the large-scale Temporal Graph Benchmark (TGB), which we adopt for evaluating \name. Recently, TPNet~\citep{tpnet} further advanced state-of-the-art link prediction by introducing temporal walk matrices with time decay, and is fully supported in \name.

\textbf{DTDG Methods.}
Discrete-time Dynamic Graph (DTDG) or snapshot-based methods represent temporal evolution as a sequence of static graph snapshots, adapting GNNs like GCN~\citep{gcn} to this setting. GCLSTM~\citep{gclstm} integrates GCNs with LSTMs~\citep{hochreiter1997long} to capture spatial and temporal dependencies, while PyG Temporal~\citep{DBLP:conf/cikm/RozemberczkiSHP21} provides a library of DTDG architectures for spatiotemporal graph learning. However, PyG Temporal lacks recent methods and standardized benchmarks like TGB. More recently, Unified Temporal Graph (UTG)~\citep{utg} demonstrated a proof-of-concept for comparing CTDG and DTDG approaches via graph discretization. While UTG offers useful insights, its implementation is slow, limited to a few datasets, and not designed for reuse. In contrast, \name supports fully vectorized graph discretization and time-iteration operations, unifying CTDG and DTDG within a single, robust framework and closing a long-standing gap in TGL.

\textbf{TGL Libraries.}
Several libraries support temporal graph learning including DyGLib~\citep{yu2023towards}, TGL~\citep{tgl_paper}, DistTGL~\citep{10.1145/3581784.3607056}, TGLite~\citep{10.1145/3620665.3640414}, and TSL~\citep{Cini_Torch_Spatiotemporal_2022}. DyGLib provides pipelines for continuous-time models but is limited by scalability, lack of modularity, and weak support for discrete-time methods~\citep{tgb2}. TGL and DistTGL offer large-scale sampling and multi-GPU execution but lack a researcher-friendly interface and have seen few recent updates. TGLite focuses on continuous-time message-flow models, while TSL addresses spatiotemporal modelling on static graphs. 

\autoref{tab:features} summarizes key aspects of these libraries. \name stands out as the only library that supports both CTDG and DTDG methods, bridging continuous- and discrete-time research paradigms. Its efficient and modular design facilitates flexible experimentation, while support for time conversion and dynamic node events enables diverse temporal graph learning tasks. Additionally, comprehensive tests and system profiling ensure reproducibility and provide research-ready infrastructure.

\begin{table}[t]
\centering
\renewcommand{\arraystretch}{1}
\definecolor{rowcolor}{HTML}{DDEBF7}
\caption{Comparison of TGL libraries. \name is the only library that meets all desirable criteria for TGL research while other libraries lack one or more criteria.}  
\label{tab:features}
\resizebox{\linewidth}{!}{%
\begin{tabular}{l|cccc|cccc}
\toprule
 & \multicolumn{4}{c|}{\textbf{TGL Features}} & \multicolumn{4}{c}{\textbf{Software Infrastructure}}\\
Library & CTDG & DTDG & Time Ops. & Node Events & Modular & Efficient & Unit Tests & Profiling\\
\midrule
\rowcolor{rowcolor} \name~(ours) & \checkmark & \checkmark & \checkmark & \checkmark & \checkmark & \checkmark & \checkmark & \checkmark\\
DyGLib & \checkmark & \texttimes & \texttimes & \texttimes  & \texttimes & \texttimes & \texttimes & \texttimes\\
TGL & \checkmark & \texttimes & \texttimes & \texttimes & \texttimes & \checkmark & \texttimes & \checkmark\\
TGLite & \checkmark & \texttimes & \texttimes & \texttimes & \checkmark & \checkmark & \checkmark & \checkmark\\
PyG Temporal & \texttimes & \checkmark & \texttimes & \texttimes & \checkmark & \checkmark & \checkmark & \texttimes\\
\bottomrule
\end{tabular}
}

\end{table}


\section{\name Framework} \label{sec:tgm_frame}

In this section, we present the formal foundations of \name. We unify continuous- and discrete-time formulations into a common representation, define graph discretization as a principled mapping from continuous events to snapshots, and introduce the hook formalism, a modular abstraction for composing graph operations. Together, these elements inform the software design in Section~\ref{sec:software}.


\textbf{\name Temporal Graph Formulation.}
First, we introduce the notation and treatment of temporal graphs in \name. On temporal graphs, events are the fundamental unit for representing the network’s evolution~\citep{DBLP:journals/jmlr/KazemiGJKSFP20}. To capture changes in graph structure and features, we distinguish between event types:

\begin{definition}[Node and Edge Events]
An edge event $(t, s, d, \vec{x}_{edge})$ is an interaction between two nodes $s$ and $d$ at time $t$ where $\vec{x}_{edge}\in \mathbb{R}^{d_{edge}}$ is the associated edge feature vector. A node event $(t, s, \vec{x}_{node})$ represents the arrival of new features $\vec{x}_{node} \in \mathbb{R}^{d_{node}}$ at node $s$ and timestamp $t$. 
\end{definition}

\definition[Temporal Graph] \label{def:tg} A temporal graph is a sequence of time-ordered events: $\mathcal{G} = \{e_0, ..., e_T\}$. Each event $e_i$ can be an edge event or a node event. Also, $\mathcal{G}$ can be associated with a static node feature matrix $\mat{X} \in \mathbb{R}^{n \times d_{static}}$ where $n$ is the number of unique nodes in $\mathcal{G}$. For any time interval $\mathcal{T} \subset \mathbb{R}^+$, the temporal sub-graph $\mathcal{G}|_{\mathcal{T}}$ contains all events in $\mathcal{G}$ intersecting $\mathcal{T}$.

\begin{figure*}[t]
  \centering
\includegraphics[width=\linewidth]{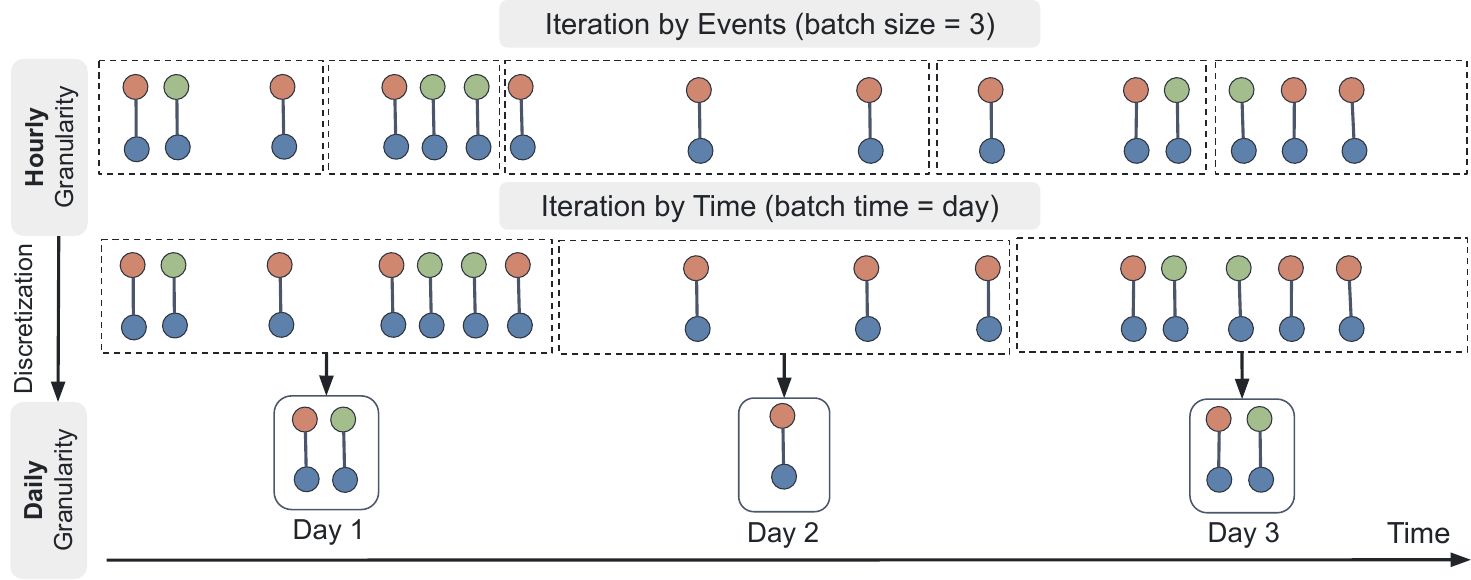}
  \caption{\name supports iteration by events and time. Discretization maps fine-grained timestamps (e.g., hourly) to coarser timestamps (e.g., daily), aggregating duplicated edges in the process.}
  \label{fig:time_ops}
\end{figure*}

\textbf{Representing Continuous-Time and Discrete-Time Graphs.} In \name, we represent temporal graphs as event sequences without distinguishing between CTDG and DTDG formulations. We argue that any temporal graph admits a native time granularity $\tau$: the coarsest unit of time (e.g., seconds) that still discriminates between all event timestamps. If real-world time is unavailable (e.g., due to privacy), \name employs a special event-ordered granularity $\tau_{\text{event}}$, preserving only the relative order of events but lacks correspondence to a real-world time granularity, thus $\tau_{\text{event}}$ is excluded from any time operations. Lastly, note that time granularities can be compared: $\hat{\tau} \leq \tau \iff \tau \textit{ is coarser than } \hat{\tau}$. This view unifies CTDG and DTDG as alternative ways of iterating over the same event stream:

\definition[CTDG: Event-based iteration] A CTDG is often expressed as a stream of events~\citep{DBLP:journals/jmlr/KazemiGJKSFP20,DBLP:conf/nips/HuangPDFHRLBRR23, tgn, DBLP:conf/kdd/YouDL22}. In \name, iterating a CTDG corresponds to using the event-ordered granularity $\tau_{\text{event}}$. Each batch contains a fixed number of events, independent of real-world time.

\definition[DTDG: Time-based iteration] A DTDG is often expressed as a sequence of static graph snapshots sampled at regularly-spaced time intervals, i.e. as $\{\mat{G}_0, \mat{G}_1, ..., \}$, where $\mat{G}_i = \{\mat{V}_i, \mat{E}_i\}$ is a static graph at snapshot $i$~\citep{utg}. 
In \name, we achieve this by iterating with a time granularity $\hat{\tau}$ that is coarser than the native graph granularity. Iterating by time produces batches $\mathcal{G}|_{[t_0, t_i]}, \mathcal{G}|_{[t_i, t_{i + 1}]}, \cdots$ where $|t_i - t_0| = |t_{i+1} - t_i| = \hat{\tau}$. 

\textbf{Discretizing Temporal Graphs.} For snapshot-based models, it is often useful to process the graph at a coarser granularity than the native $\tau$ (e.g., daily instead of second-wise). Discretization converts the underlying network to this coarser timeline by collapsing duplicate edges within each time interval:

\definition[Time Granularity Discretization.] Let $\mathcal{G}$ be a temporal graph with native time granularity $\tau$. For any $\hat{\tau} \geq \tau$, the discretization operator:
\begin{align}
    \mathcal{\psi}_{r}: (\mathcal{G}, \tau) \mapsto (\hat{\mathcal{G}}, \hat{\tau})
\end{align}
maps $\mathcal{G}$ to coarser granularity $\hat{\tau}$, groups events into equivalence classes induced by $\hat{\tau}$ and applies a reduction operator $r$ to each class. The resulting graph 
$\hat{\mathcal{G}}$ contains one representative event per class. Figure~\ref{fig:time_ops} illustrates these time operations in \name.

\textbf{\name Learning Tasks}
The common goal in TG is to forecast the structure or property of the graph in the future. \name supports ML on all levels of the graph, namely link, node and graph tasks:
\begin{itemize}[leftmargin=*, noitemsep, topsep=0.2pt]
\item \textbf{Dynamic Link Property Prediction.} Given the temporal sub-graph $\mathcal{G}|_{[t_0, t_i]}$, predict some property (or existence) of a link between a node pair $(s,d)$ at a future timestamp $t$ where $t > t_i$.  
\item \textbf{Dynamic Node Property Prediction.} Given the temporal sub-graph $\mathcal{G}|_{[t_0, t_i]}$, predict some property of a node $s$ at a future timestamp $t$ where $t > t_i$.  
\item \textbf{Dynamic Graph Property Prediction.} Given the temporal sub-graph $\mathcal{G}|_{[t_0, t_i]}$, predict some property of $\mathcal{G}|_{[t', t'']}$ over a future time interval $[t', t'']$ where $t_i < t'\leq t''$.
\end{itemize}



\textbf{\name Hooks and Recipes.} We formalize a TGL workflow as a composition of transformations called hooks. Each hook specifies a typed contract on batch attributes, and recipes are valid precisely when their signatures compose. 

\begin{definition}[Materialized Batch] Let $G|_{\mathcal{T}}$ be a temporal subgraph. We denote by $B|_{\mathcal{T}, \mathcal{A}}$ the materialized batch associated with a set of properties $\mathcal{A}$. Intuitively, $\mathcal{A}$ captures the attributes that enrich the slice of data, typically tensors required by a model (e.g. neighborhood information in message-passing architectures). 
\end{definition}

\begin{definition}[Hook] A hook $\phi_{\mathcal{R}, \mathcal{P}}$ is a transformation on a materialized batch:
\begin{align}
    \phi_{\mathcal{R}, \mathcal{P}}: \mathcal{B}|_{\mathcal{T}, \mathcal{A}} \mapsto \mathcal{B}|_{\mathcal{T}, \mathcal{A} \cup \mathcal{P}}
\end{align}
which declares a contract based on the attributes required on the input $\mathcal{R} \subset \mathcal{A}$, and the attributes produced $\mathcal{P}$, so that the batch transformed by $\phi$ has attributes $\mathcal{A} \cup \mathcal{P}$. Table \ref{tab:hooks} illustrates several common temporal graph operations expressed as hooks using the notation introduced here.
\end{definition}

The real power of hooks is unlocked by composing their transformations to express complete temporal graph workflows. The notion of a hook recipe formalizes this.

\begin{figure*}[t]
  \centering
  \includegraphics[width=0.85\linewidth]{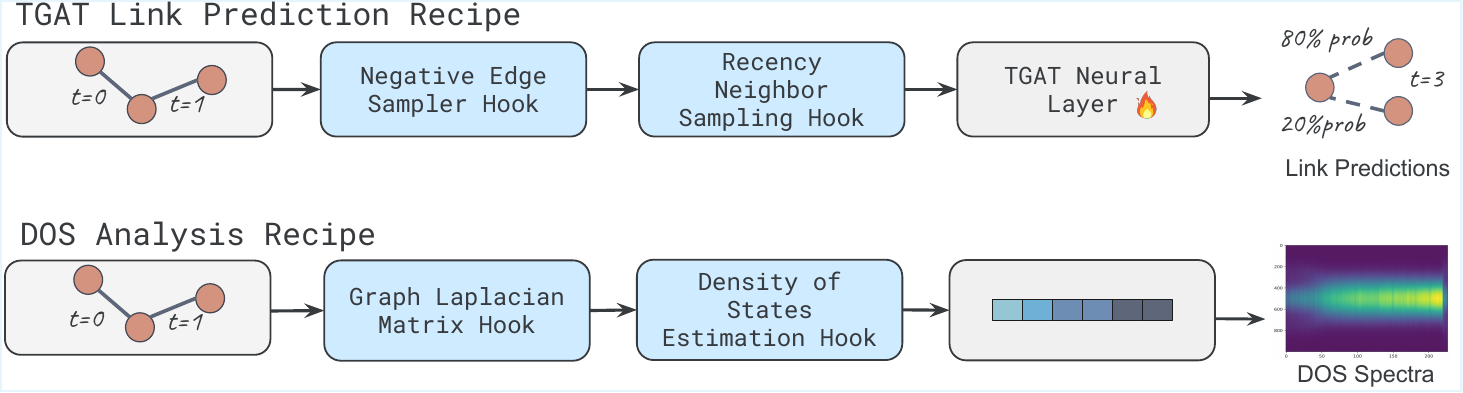}
  \caption{Example recipes in \name: TGAT link prediction and Density of States Analysis. \name provides a unified ecosystem supporting both representation learning and temporal graph analytics. The constituent hooks are modular, enabling reuse across different workflows within the community.}
  \label{fig:recipe}
\end{figure*}

\begin{definition}[Hook Recipe] A set of hooks $\{\phi^1_{\mathcal{R}_1, \mathcal{P}_1}, ..., \phi^k_{\mathcal{R}_k, \mathcal{P}_k}\}$ induces an ordering given by their dependencies:
\begin{align}
    \phi^i \to  \phi^j \iff \mathcal{P}_i\cap \mathcal{R}_j \neq \emptyset
\end{align}

We call this a hook recipe if this dependency graph is acyclic and every required is satisfied, i.e. $\forall j, \mathcal{R}_j \subset \bigcup_{i < j} \mathcal{P}_i$. Thus, any hook recipe admits a valid ordering by topological sort.
With this framework, exploring new research is simpler as complex workflows can be expressed with minimal boilerplate. Figure~\ref{fig:recipe} illustrates how ML and analytics pipelines are represented as recipes in \name.
\end{definition}

\begin{table}[t]
\centering
\renewcommand{\arraystretch}{1}
\caption{Examples of common temporal graph operations represented as hooks, and their attributes.}
\label{tab:hooks}
\resizebox{0.9\linewidth}{!}{%
\begin{tabular}{l|ccccc}
\toprule
\textbf{Hook Type} 
& \multicolumn{2}{c}{\textbf{Neighbor Sampling}} 
& \textbf{Evaluation} 
& \multicolumn{1}{c}{\textbf{Device Ops.}} 
& \multicolumn{1}{c}{\textbf{Analytics}} \\
& Recency
& Uniform
& TGB Eval
& GPU Transfer
& DOS Estimate \\
\midrule
$\mathcal{R}$ (Requires) & $\{\text{negatives}\}$ & $\{\text{negatives}\}$ & $\emptyset$ & $\emptyset$ & $\emptyset$ \\
$\mathcal{P}$ (Produces)  & $\{\text{neighbors}\}$ & $\{\text{neighbors}\}$ & $\{\text{negatives}\}$ & $\emptyset$ & $\{\text{DOS}\}$ \\
\midrule
\end{tabular}
}
\end{table}

\section{\name Software Library} \label{sec:software}

\begin{figure*}[t]
  \centering
  \includegraphics[width=0.9\linewidth]{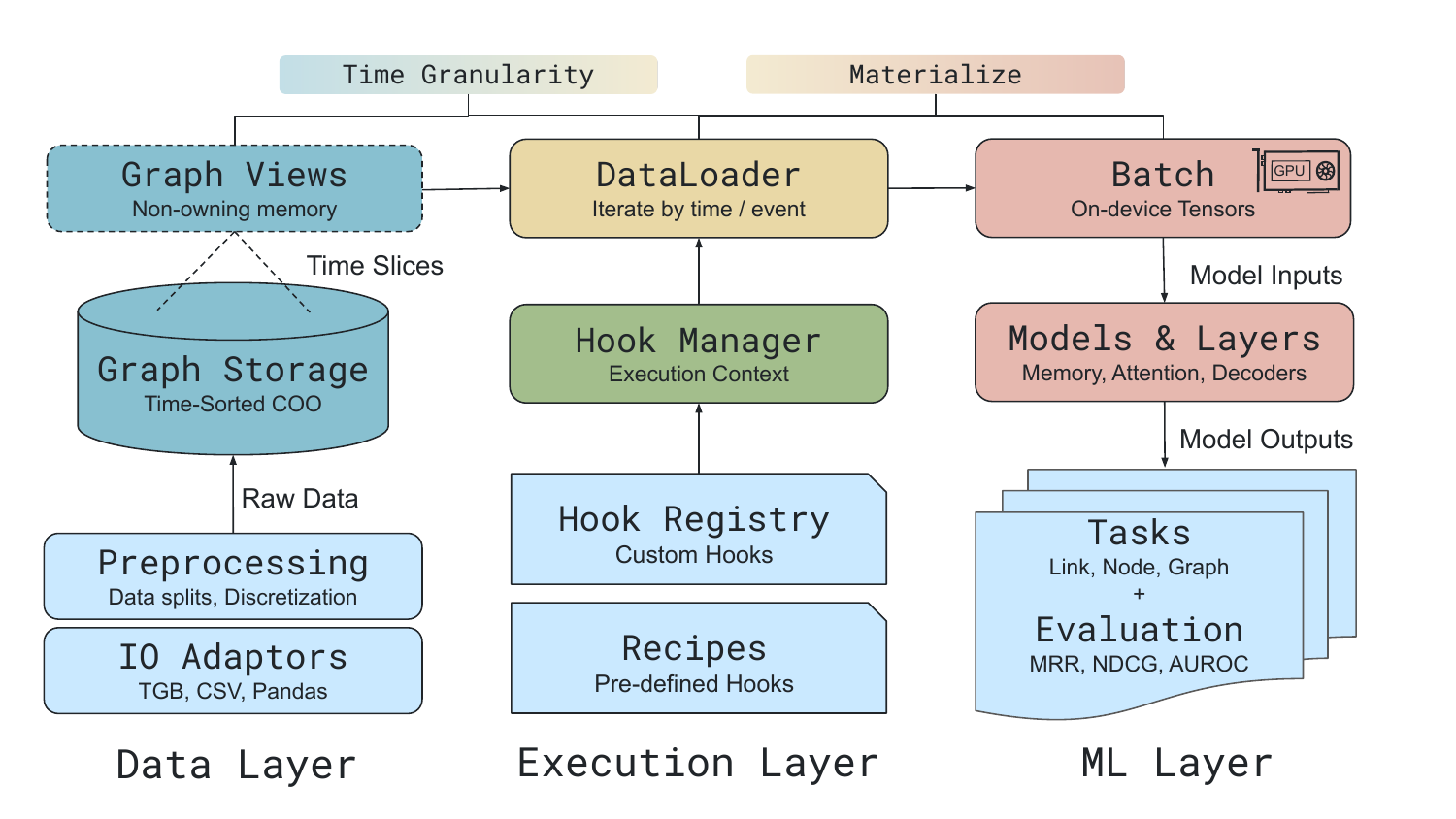}
  \caption{Three Layer Architecture of \name: data layer (left), with IO adaptors and preprocessing, immutable COO graph storage, and lightweight sub-graph views; execution layer (middle), where users register custom hooks or apply pre-built recipes through the hook manager and dataloader to inject execution logic; and ML layer (right), where batches are materialized on device and used for node-, link-, or graph-level prediction. Light blue elements denote user-facing APIs.}
  \label{fig:arch}

\end{figure*}

We now describe the software implementation that realizes the framework described in the previous section. Figure~\ref{fig:arch} presents the high-level system design: the data layer is an immutable time-sorted coordinate format (COO) storage with lightweight graph views for efficient slicing; the execution layer is built around a hook manager that transparently performs complex transformations (e.g., temporal neighbors); and the ML layer materializes batches on-device for model computation. This separation of concerns yields workflows that are efficient and extensible, as we show in Section~\ref{sec:research_exp}.

\textbf{IO Adaptors and Data Preprocessing.} \name streamlines experimentation by integrating the widely-used benchmark dataset: TGB~\citep{huang2023temporal,tgb2}, in the form of IO Adapters, including loading, preprocessing, and train/validation/test splits. This allows researchers to start experiments immediately and compare models consistently with minimal overhead. Custom adapters are also supported via CSV and Pandas. Our design makes it straightforward to incorporate new benchmarks while ensuring consistent evaluation across all datasets (see Appendix~\ref{app:data}).

\textbf{Graph Storage and Graph Views.} The storage exposes an interface for graph queries, implemented using a time-sorted COO with a cached index. This enables binary search over timestamps, which is critical for recent-neighbor retrieval. The backend is designed for extension, allowing alternative layouts~\citep{lpma, gpma} so future models can use the most efficient data structures for their workload. Backed by the storage, graph views provide lightweight, concurrency-safe access to temporal sub-graphs. Each view tracks time boundaries and encodes read-access through the time granularity abstraction. This enables \name to perform both CTDG and DTDG-style loading, making it straightforward to study the effects of snapshot resolutions, as illustrated in Section~\ref{sec:research_exp}. Our discretization is fully vectorized, enabling efficient snapshot creation, as demonstrated in Table~\ref{tab:discretization_latency}.

\textbf{Hook Registry and Management.}
Building on the graph abstractions, hooks are transformations that can be combined to create complex workflows (see Section s~\ref{sec:tgm_frame}). The HookManager handles shared state, resolves dependencies, and executes transparently during data loading. A key-value interface allows hooks to be registered under specific conditions (e.g., analytics hooks, training hooks). We provide pre-defined recipes for common tasks such as TGB link prediction, helping new practitioners avoid common pitfalls like mismanaging state across data splits or using incorrect negatives.

\textbf{Diverse Model and Task Support.}
\name provides PyTorch modules tailored for TGL, including memory units, attention layers, and link decoders. With this, \name implements a range of TG methods, from baselines like EdgeBank~\citep{edgebank}, to message passing-based models like TGAT~\citep{tgat}, and frontier models like DyGFormer~\citep{dygfomer} and TPNet~\citep{tpnet}. Crucially, learnable components are decoupled from graph management, making it easy for researchers to prototype new models.

\begin{figure*}[!ht]
  \centering
  \includegraphics[width=0.9\linewidth]{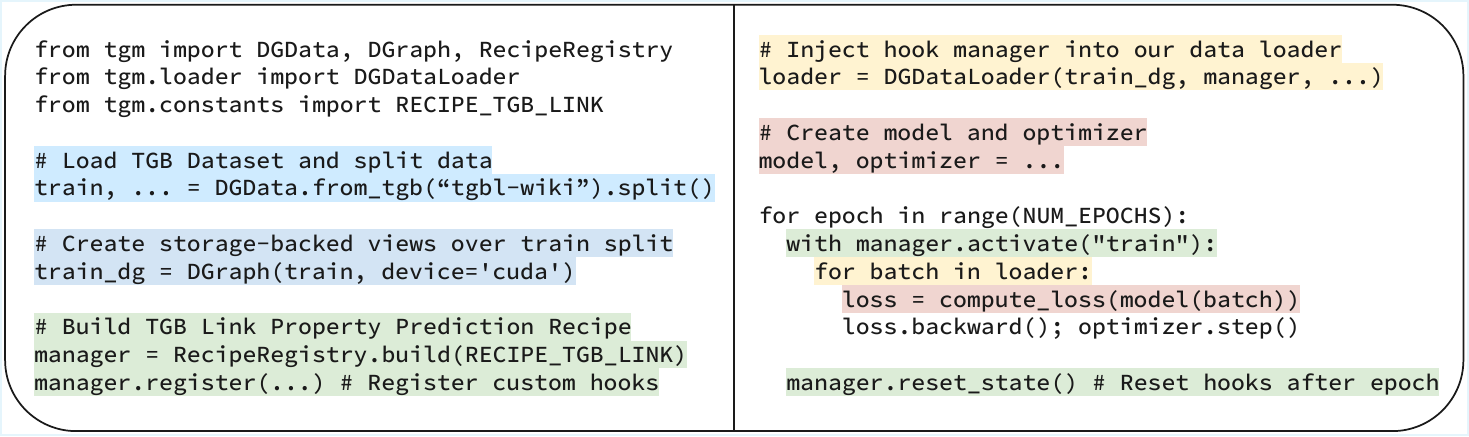}
  \caption{Example workflow in \name. Left: dataset loading, graph creation, and hook registration; Right: manager injection, model setup, and training loop with automatic hook activation. Highlighted code maps to system components from Figure~\ref{fig:arch}.}
  \label{fig:workflow}
\end{figure*}

\textbf{Streamlined TGL Workflows.} Figure~\ref{fig:workflow} provides a high-level overview of a typical workflow in \name, illustrating how data preparation, graph creation, hook registration, and model training are orchestrated. Loading a temporal graph is straightforward, and hook registration can be shared and reused across different workflows, enabling code reuse. Registered hooks dynamically inject behaviour during data loading, ensuring models automatically receive the appropriate tensors. This unifies the model interface and explicitly defines which batch attributes each model consumes. The manager’s reset method exposes a single API for clearing the state of all active hooks. More complex workflows can be implemented by registering hooks under key-value pairs.

\textbf{Robust and Research-Ready Infrastructure.}
Finally, \name is built following modern software engineering practices to ensure reliability, maintainability, and ease of use. We use type hinting throughout the codebase, which unifies model APIs and improve usability. Continuous integration pipelines run end-to-end tests on all layers, hooks, and graph APIs with test coverage to ensure correctness. Performance monitoring utilities can track GPU usage with support for tools such as FlameProf~\citep{flame} to help identify bottlenecks. We also provide detailed tutorials, documentation, and examples for link, node and graph tasks. Overall, TGM provides a high-quality, research-ready platform that lowers the barrier to TG research while supporting efficient experimentation.


\section{Experiments} \label{sec:exp}

In this section, we evaluate \name along two dimensions central to its design: efficiency and research extensibility. Correctness results are deferred to Appendix~\ref{app:correctness}, where we show that \name faithfully reproduces prior library performance. The appendix also includes peak memory measurements (~\ref{sec:gpu}) and a detailed runtime breakdown (~\ref{sec:profiler}) collected with \name’s profiling tools.

\subsection{Efficiency Benchmark} \label{sub:efficiency}

We evaluate \name on two standard TGL tasks: dynamic link property prediction and dynamic node property prediction. Because graph discretization is a core operation in DTDG methods, we additionally benchmark the efficiency of \name in this setting. All datasets are stored in CPU host memory and transferred to GPU when required. Full experimental details, including model hyperparameters and compute resources, are provided in Appendix~\ref{app:details}.

\begin{table*}[t]
\centering
\caption{Training time per epoch (seconds, ↓) for link property prediction. The \hfirst \text{} and \hsecond \text{} best results are highlighted (\cna \text{} marks unsupported). \name achieves competitive performance to the system-optimized TGLite library on TGAT and TGN models while supporting a broader range of architectures, and consistently outperforms the widely used research library DyGLib across all datasets and models, delivering a $4.4\times$ speedup on the transformer-based DyGFormer architecture.}
\label{tab:link_prop_perf}

\resizebox{\linewidth}{!}{%
\begin{tabular}{l|rrrr|rrrr|rrrr}
\toprule
\multirow{2}{*}{Model} 
& \multicolumn{4}{c|}{\texttt{Wikipedia}} 
& \multicolumn{4}{c|}{\texttt{Reddit}} 
& \multicolumn{4}{c}{\texttt{LastFM}} \\
& \name & DyGLib & TGLite & TGL 
& \name & DyGLib & TGLite & TGL  
& \name & DyGLib & TGLite & TGL  \\
\midrule
TGAT       & \csecond{6.97} & 41.24 & \cfirst{4.85} & 10.00 & \csecond{28.23} & 182.21 & \cfirst{25.00} & 53.25 & \csecond{55.32} & 349.31 & \cfirst{38.00} & 85.12 \\
TGN        & \csecond{10.59} & 63.37 & \cfirst{6.80} & 23.32 & \csecond{61.25} & 287.06 & \cfirst{60.50} & 125.23 & \cfirst{91.23} & 392.98 & \csecond{92.93} & 250.00 \\
DyGFormer  & \cfirst{17.00} & \csecond{75.10} & \multicolumn{1}{c}{\cna} & \multicolumn{1}{c|}{\cna} & \cfirst{72.29} & \csecond{326.60} & \multicolumn{1}{c}{\cna} & \multicolumn{1}{c|}{\cna} & \cfirst{142.40} & \csecond{633.99} & \multicolumn{1}{c}{\cna} & \multicolumn{1}{c}{\cna} \\
TPNet      & \cfirst{12.28} & \multicolumn{1}{c}{\cna} & \multicolumn{1}{c}{\cna} & \multicolumn{1}{c|}{\cna} & \cfirst{49.79} & \multicolumn{1}{c}{\cna} & \multicolumn{1}{c}{\cna} & \multicolumn{1}{c|}{\cna} & \cfirst{97.23} & \multicolumn{1}{c}{\cna} & \multicolumn{1}{c}{\cna} & \multicolumn{1}{c}{\cna} \\
GCLSTM     & \cfirst{3.56} & \multicolumn{1}{c}{\cna} & \multicolumn{1}{c}{\cna}  & \multicolumn{1}{c|}{\cna}  & \cfirst{9.17} & \multicolumn{1}{c}{\cna}  & \multicolumn{1}{c}{\cna}  & \multicolumn{1}{c|}{\cna}  & \cfirst{140.69} & \multicolumn{1}{c}{\cna}  & \multicolumn{1}{c}{\cna}  & \multicolumn{1}{c}{\cna} \\
GCN        & \cfirst{2.50} & \multicolumn{1}{c}{\cna}  & \multicolumn{1}{c}{\cna}  & \multicolumn{1}{c|}{\cna}  & \cfirst{7.88} & \multicolumn{1}{c}{\cna}  & \multicolumn{1}{c}{\cna}  & \multicolumn{1}{c|}{\cna}  & \cfirst{96.89} & \multicolumn{1}{c}{\cna}  & \multicolumn{1}{c}{\cna}  & \multicolumn{1}{c}{\cna} \\
\bottomrule
\end{tabular}
}
\end{table*}

\textbf{Link Property Prediction.} We benchmark \name against state-of-the-art libraries on the dynamic link property prediction task using three standard datasets: \texttt{Wikipedia}, \texttt{Reddit}, and \texttt{LastFM}~\footnote{Referred to as tgbl-wiki, tgbl-subreddit, and tgbl-lastfm in TGB.}. Competing baselines include DyGLib~\citep{dygfomer}, TGL~\citep{tgl_paper}, and TGLite~\citep{10.1145/3620665.3640414}, all of which are designed primarily for continuous-time models.
Table~\ref{tab:link_prop_perf} reports training time per epoch across models implemented in \name and competing libraries. First, \name uniquely supports the widest range of architectures, spanning both CTDG and DTDG methods. In particular, DTDG models such as GCLSTM and GCN are supported via graph discretization and iterate-by-time functionality, and \name is the only library with native support for TPNet~\citep{DBLP:conf/nips/LuSZL24}, the state-of-the-art link prediction model on TGB as of September 2025. Second, \name consistently ranks among the top two fastest implementations across datasets and models. It outperforms DyGLib and TGL in all cases, and is only slightly behind the highly specialized TGLite on TGAT and TGN. For example, \name achieves a $4.4\times$ speedup over the alternative DyGFormer implementation on \texttt{Wikipedia}. A key driver of performance is our fully vectorized recency sampler, implemented with a circular buffer in PyTorch-native code, which enables cache-friendly memory access. Finally, \name offers native support for TGB evaluation, the standard benchmark protocol. Appendix~\ref{app:more_perf} shows that \name can be up to $246\times$ than DyGLib for TGN on \texttt{Wikipedia}, owing to batch-level de-duplication and efficient data handling: while DyGLib repeatedly samples neighbors for each prediction, \name samples once per batch. By contrast, TGL and TGLite do not support this one-vs-many evaluation, limiting their benchmarking robustness compared to \name.

\noindent
\begin{minipage}[t]{0.4\textwidth}  
\textbf{Node Property Prediction.} We benchmark \name on the dynamic node property prediction task, comparing against both DyGLib and the native TGB implementations on the \texttt{Trade} and \texttt{Genre} datasets. TGL and TGLite do not support this task. Table~\ref{tab:node_prop_perf} reports training time per epoch. Compared to DyGLib, \name achieves up to a $10\times$ speedup for TGN on \texttt{Trade} while reducing training time by 80 seconds on \texttt{Genre}. Moreover, \name is the only library supporting node property prediction for DTDG models: GCLSTM, GCN, and TGCN. Note, we encountered an \small{\texttt{OOM}} while running DyGLib on \texttt{Genre} with our 64GB RAM allocation (see Appendix~\ref{app:details}), requiring 256GB of memory to produce the results reported in Table~\ref{tab:node_prop_perf}.
\end{minipage}%
\hfill
\begin{minipage}[t]{0.57\textwidth}  
\centering

\captionof{table}{Training time per epoch (seconds, ↓) for dynamic node property prediction. The \hfirst \text{} and \hsecond \text{} best results are highlighted (\cna \text{} marks unsupported). \name has the best all-around performance and uniquely supports message-passing (TGN), snapshots-based (e.g. TGCN), and transformer-based (DyGFormer) models.}
\label{tab:node_prop_perf}
\resizebox{\linewidth}{!}{%
\begin{tabular}{l|rrr|rrr}
\toprule
\multirow{2}{*}{Model} & \multicolumn{3}{c|}{\texttt{Trade}} & \multicolumn{3}{c}{\texttt{Genre}} \\
 & \name & DyGLib & TGB & \name & DyGLib & TGB \\
\midrule
TGN        & \csecond{12.94} & 19.37 & \cfirst{11.07} & \cfirst{208.88} & 918.46 & \csecond{281.36} \\
DyGFormer  & \cfirst{16.24} & \csecond{117.13} & \cna & \cfirst{70.89} & \csecond{3539.95} & \cna \\
P.F.       & \cfirst{0.41} & 2.09 & \csecond{0.78} & \csecond{38.15} & 41.73 & \cfirst{35.58} \\
TGCN       & \cfirst{0.85} & \cna & \cna & \cfirst{17.27} & \cna & \cna \\
GCLSTM     & \cfirst{0.88} & \cna & \cna & \cfirst{17.71} & \cna & \cna \\
GCN        & \cfirst{0.80} & \cna & \cna & \cfirst{17.21} & \cna & \cna \\
\bottomrule
\end{tabular}
}
\end{minipage}


\begin{minipage}[t]{0.56\textwidth}  
\textbf{Graph Discretization.} Enabling DTDG models on CTDG tasks requires discretizing the original graph into snapshots. We compare \name's implementation with that of UTG~\citep{utg}. Table~\ref{tab:discretization_latency} shows that \name achieves dramatic speedups, up to $433\times$ on \texttt{LastFM}. This improvement stems from a fully vectorized, PyTorch-native implementation that avoids cache-unfriendly Python dictionaries and other overheads common in prior repositories. This result underscores our commitment to high-performance, research-ready tooling, setting \name apart from existing libraries in efficiency and engineering standards.

\end{minipage}%
\hfill
\begin{minipage}[t]{0.4\textwidth}  
\centering
\captionof{table}{Discretization Latency to Hourly Snapshots (seconds, ↓). \name has substantial speedups due to our vectorized, PyTorch-native implementation.}
\label{tab:discretization_latency}
\resizebox{\linewidth}{!}{%
\begin{tabular}{l | rrr}
\toprule
Dataset & UTG & \name & Speedup \\
\midrule
\texttt{Wikipedia} & 1.94 & 0.04 & 49.62× \\
\texttt{Reddit}    & 8.83 & 0.21 & 41.63× \\
\texttt{LastFM}    & 19.94 & 0.05 & 433.39× \\
\bottomrule
\end{tabular}
}
\end{minipage}


\begin{table}[t]
\centering
\caption{The choice of snapshot time granularity significantly affects link prediction performance. Reported metric is MRR (↑) with the \hfirst{} and \hsecond{} best result for each dataset highlighted.}
\label{tab:snapshot_granularity_effect}
\small
\setlength{\tabcolsep}{4pt}
\renewcommand{\arraystretch}{0.9}
\resizebox{0.9\linewidth}{!}{%
\begin{tabular}{l|ccc|ccc}
\toprule
\multirow{2}{*}{Time Gran.} & \multicolumn{3}{c|}{\texttt{Wikipedia}} & \multicolumn{3}{c}{\texttt{Reddit}} \\
 & GCN & T-GCN & GCLSTM & GCN & T-GCN & GCLSTM \\
\midrule
Hourly & 0.510 \scriptsize{$\pm$ 0.001} & 0.509 \scriptsize{$\pm$ 0.004} & 0.395 \scriptsize{$\pm$ 0.022} & \cfirst{0.529 \scriptsize{$\pm$ 0.012}} & \csecond{0.374 \scriptsize{$\pm$ 0.004}} & 0.219 \scriptsize{$\pm$ 0.003} \\
Daily  & \cfirst{0.702 \scriptsize{$\pm$ 0.007}} & \csecond{0.540 \scriptsize{$\pm$ 0.008}} & 0.372 \scriptsize{$\pm$ 0.017} & 0.266 \scriptsize{$\pm$ 0.007} & 0.231 \scriptsize{$\pm$ 0.003} & 0.212 \scriptsize{$\pm$ 0.004} \\
Weekly & 0.393 \scriptsize{$\pm$ 0.005} & 0.330 \scriptsize{$\pm$ 0.009} & 0.323 \scriptsize{$\pm$ 0.010} & 0.191 \scriptsize{$\pm$ 0.002} & 0.212 \scriptsize{$\pm$ 0.001} & 0.206 \scriptsize{$\pm$ 0.004} \\
\bottomrule
\end{tabular}%
}
\end{table}

\subsection{\name Research Experiments}
\label{sec:research_exp}

In addition to its efficiency, \name is designed as a flexible framework for exploring research questions in temporal graph learning. By supporting both CTDG and DTDG methods, along with native time conversions and composable hooks, \name allows researchers to implement and test novel ideas effortlessly. We ran all three example experiments using a single script, which we include in our anonymized code release. These experiments investigate the following questions: RQ1: How accurately can we predict the future evolution of a graph property? RQ2: How does the time granularity of graph snapshots impact DTDG performance on a continuous-time graph? RQ3: How do batching strategies, by fixed edges versus by time, affect the performance of a CTDG model?

\begin{minipage}[t]{0.51\textwidth}
\textbf{RQ1: Dynamic Graph Property Prediction.} 
Graph-level tasks require grouping edges into snapshots. The ability to natively support iteration by time is unique to \name framework, thus allowing researchers to effortlessly explore research questions in dynamic graph property prediction. As shown in Table~\ref{tab:graphproppred}, we leverage this capability to benchmark models on predicting whether a future transaction network snapshot will grow, a key problem for understanding network evolution. The results highlight the sensitivity of model performance to temporal granularity: T-GCN performs best on weekly snapshots with an AUC of 0.800, while GCLSTM excels at the daily scale with an AUC of 0.589.
\end{minipage}
\hfill
\begin{minipage}[t]{0.45\textwidth}
\centering
\captionof{table}{Binary classification task predicting whether the next daily snapshot will see an increase in the number of edges. Reported metric is AUC (↑) with the \hfirst{} and \hsecond{} best result for each dataset highlighted. }
\label{tab:graphproppred}
\resizebox{\linewidth}{!}{%
\begin{tabular}{l|r|r }
\toprule
Model & \texttt{Wikipedia} & \texttt{Reddit}  \\
\midrule
 P.F. & 0.018 \scriptsize{$\pm$ 0.058} & \cfirst{0.617 \scriptsize{$\pm$ 0.047}} \\
 T-GCN & \cfirst{\textbf{0.667 \scriptsize{$\pm$ 0.083}}}& \csecond{0.600 \scriptsize{$\pm$ 0.147}}\\
 GCLSTM& 0.567 \scriptsize{$\pm$ 0.047}  & 0.526 \scriptsize{$\pm$ 0.020} \\
GCN & \csecond{0.577 \scriptsize{$\pm$ 0.053}}  & 0.200 \scriptsize{$\pm$ 0.000} \\

\bottomrule
\end{tabular}
}
\end{minipage}


\textbf{RQ2: Effect of Time Granularity for DTDG methods. }
Table~\ref{tab:snapshot_granularity_effect} demonstrates that the choice of snapshot granularity, i.e. hourly, daily, or weekly, has a substantial impact on the performance of snapshot-based temporal graph models. On the Wikipedia dataset, the impact is particularly pronounced: GCN’s MRR increases by 30\% when moving from weekly to daily snapshots, while T-GCN and GCLSTM improve by 21\% and 5\%, respectively. On Reddit, the same trend is observed, though less extreme: GCN achieves 0.529 MRR with hourly snapshots, dropping to 0.191 with weekly snapshots. These results underscore the importance of selecting an appropriate snapshot granularity for DTDG models. \name makes this process effortless, allowing users to adjust the time granularity with a single line of code, treating it effectively as a hyperparameter.

\begin{minipage}[t]{0.52\textwidth}
\textbf{RQ3: Effect of Batch Size for CTDG methods.} Our analysis reveals that the configuration of the evaluation process itself is a critical, yet previously overlooked, hyperparameter in temporal graph learning. As demonstrated in Table~\ref{tab:batch_size}, the choice of validation batch size and temporal unit significantly impacts the reported performance of the TGAT model on link prediction.  Note that when iterating by time, the number of edges in each batch is different, however, each batch spans a fixed amount of time instead. We observe a pronounced degradation in MRR with larger batch sizes and coarser temporal units (e.g., Day versus Hour). \name supports flexible temporal batching via our graph formulation, enabling the investigation of batch size at test time. 

\end{minipage}
\hfill
\begin{minipage}[t]{0.43\textwidth}
\centering
\captionof{table}{The choice of validation batch size and batch unit affects the performance of TGAT link prediction on \texttt{Wikipedia} dataset. \hfirst{} and \hsecond{} are highlighted.}
\label{tab:batch_size}
\small
\begin{tabular}{l| c |c}
\toprule
 & Size/Unit & Test MRR (↑)\\
 \cmidrule(lr){2-2} \cmidrule(lr){3-3} 
 \multirow{3}{*}{Batch size} & 1 & \cfirst{0.449 $\pm$ \scriptsize{0.001}}\\
 & 50 & 0.414 $\pm$ \scriptsize{0.006} \\
 & 100 & \csecond{0.414 $\pm$ \scriptsize{0.004}} \\
 & 200 & 0.403 $\pm$ \scriptsize{0.004} \\
\cmidrule(lr){1-1} \cmidrule(lr){2-2} \cmidrule(lr){3-3} 
 \multirow{2}{*}{Batch unit} & Hour & 0.402 $\pm$ \scriptsize{0.012} \\
 & Day & 0.349 $\pm$ \scriptsize{0.004} \\
\bottomrule
\end{tabular}
\end{minipage}

\section{Conclusion}

We present \name, a modular and efficient framework for temporal graph learning built around a novel hook formalism. By decoupling graph operations from model logic, \name enables rapid prototyping and code reuse, unifying CTDG and DTDG methods under a single research-ready library. Efficiency-wise, \name is highly competitive and on average $7.8\times$ faster in training than the widely used DyGLib. 
We ultimately envision \name as a foundation for a shared ecosystem where models, hooks, and analytics can be seamlessly composed and reused, accelerating TGL research.



\section*{Reproducibility}

The \name code repo is publicly available at \url{https://github.com/tgm-team/tgm}. The package is also available via PyPI install at \url{https://pypi.org/project/tgm-lib/}. All experiments use fixed random seeds, and full hyperparameters are listed in Table~\ref{tab:hyperparams}. The Python environment was built deterministically and managed with the \texttt{uv} package manager. Benchmarks were conducted in isolated SLURM jobs, and additional details on compute resources and experimental setup is provided in Appendix~\ref{app:details}.
\subsubsection*{Acknowledgments}
This research was supported by the Canadian Institute for Advanced Research (CIFAR AI chair program) and the AI Security Institute (AISI) grant: Towards Trustworthy AI Agents for Information Veracity and the EPSRC Turing AI World-Leading Research Fellowship No. EP/X040062/1 and EPSRC AI Hub No. EP/Y028872/1. Shenyang Huang was supported by the Natural Sciences and Engineering Research Council of Canada (NSERC) Postgraduate Scholarship Doctoral (PGS D) Award and Fonds de recherche du Québec - Nature et Technologies (FRQNT) Doctoral Award. This research was also enabled in part by compute resources provided by Mila (mila.quebec).

\bibliographystyle{iclr2026_conference}
\bibliography{ref}

\newpage
\appendix



\section{Additional Results}\label{app:more_perf}

\subsection{Validation Latency Benchmarks}

In Table~\ref{tab:eval}, we report the TGB validation evaluation time per epoch for \name and other libraries. Note that \name supports highly optimized evaluation time for the robust TGB link prediction evaluation when compared to DyGLib. \name consistently outperforms the widely used research library DyGLib across datasets and models. TGLite and TGL do not support the one-vs-many TGB-based link prediction evaluation ~\citep{tgb2}.

\begin{table*}[t]
\centering
\caption{Validation time per epoch (seconds, ↓) for link property prediction (top) and node property prediction (bottom). The \hfirst \text{} and \hsecond \text{} best results are highlighted (\cna \text{} marks unsupported).  \normalsize{\texttt{OOT}} indicates that a single validation epoch did not complete after 3 hours.}

\resizebox{\linewidth}{!}{
\begin{tabular}{l|rrrr|rrrr|rrrr}
\toprule
\multirow{2}{*}{Model} 
& \multicolumn{4}{c|}{\texttt{Wikipedia}} 
& \multicolumn{4}{c|}{\texttt{Reddit}} 
& \multicolumn{4}{c}{\texttt{LastFM}} \\
& \name & DyGLib & TGLite & TGL 
& \name & DyGLib & TGLite & TGL  
& \name & DyGLib & TGLite & TGL  \\
\midrule
EdgeBank       & \cfirst{11.08} & \csecond{950.05} & \multicolumn{1}{c}{\cna} & \multicolumn{1}{c|}{\cna} & \cfirst{50.01} & \csecond{134.55} & \multicolumn{1}{c}{\cna} & \multicolumn{1}{c|}{\cna} & \cfirst{223.01} & \csecond{470.08} & \multicolumn{1}{c}{\cna} & \multicolumn{1}{c}{\cna} \\
TGAT       & \cfirst{532.89} & \csecond{2898.53} & \multicolumn{1}{c}{\cna} & \multicolumn{1}{c|}{\cna} & \cfirst{2241.70} & \multicolumn{1}{c}{\normalsize{\texttt{OOT}}} & \multicolumn{1}{c}{\cna} & \multicolumn{1}{c|}{\cna} & \cfirst{4163.20} & \multicolumn{1}{c}{\normalsize{\texttt{OOT}}} & \multicolumn{1}{c}{\cna} & \multicolumn{1}{c}{\cna} \\
TGN        & \cfirst{13.84} & \csecond{3404.82} & \multicolumn{1}{c}{\cna} & \multicolumn{1}{c|}{\cna} & \cfirst{60.30} & \multicolumn{1}{c}{\normalsize{\texttt{OOT}}} & \multicolumn{1}{c}{\cna} & \multicolumn{1}{c|}{\cna} & \cfirst{112.23} & \multicolumn{1}{c}{\normalsize{\texttt{OOT}}} & \multicolumn{1}{c}{\cna} & \multicolumn{1}{c}{\cna} \\
DyGFormer  & \cfirst{6.97} & \csecond{6125.05} & \multicolumn{1}{c}{\cna} & \multicolumn{1}{c|}{\cna} & \cfirst{1856.78} & \multicolumn{1}{c}{\normalsize{\texttt{OOT}}} & \multicolumn{1}{c}{\cna} & \multicolumn{1}{c|}{\cna} & \cfirst{3554.252} & \multicolumn{1}{c}{\texttt{OOT}} & \multicolumn{1}{c}{\cna} & \multicolumn{1}{c}{\cna} \\
TPNet      & \cfirst{408.91} & \multicolumn{1}{c}{\cna} & \multicolumn{1}{c}{\cna} & \multicolumn{1}{c|}{\cna} & \cfirst{1735.71} & \multicolumn{1}{c}{\cna} & \multicolumn{1}{c}{\cna} & \multicolumn{1}{c|}{\cna} & \cfirst{3308.91} & \multicolumn{1}{c}{\cna} & \multicolumn{1}{c}{\cna} & \multicolumn{1}{c}{\cna} \\
GCLSTM     & \cfirst{11.92} & \multicolumn{1}{c}{\cna} & \multicolumn{1}{c}{\cna} & \multicolumn{1}{c|}{\cna} & \cfirst{51.68} & \multicolumn{1}{c}{\cna} & \multicolumn{1}{c}{\cna} & \multicolumn{1}{c|}{\cna} & \cfirst{110.16} & \multicolumn{1}{c}{\cna} & \multicolumn{1}{c}{\cna} & \multicolumn{1}{c}{\cna} \\
GCN        & \cfirst{11.70} & \multicolumn{1}{c}{\cna} & \multicolumn{1}{c}{\cna} & \multicolumn{1}{c|}{\cna} & \cfirst{50.88} & \multicolumn{1}{c}{\cna} & \multicolumn{1}{c}{\cna} & \multicolumn{1}{c|}{\cna} & \cfirst{102.56} & \multicolumn{1}{c}{\cna} & \multicolumn{1}{c}{\cna} & \multicolumn{1}{c}{\cna} \\
\bottomrule
\end{tabular}
}

\vspace{1em} 

\resizebox{0.55\linewidth}{!}{
\begin{tabular}{l|rrr|rrr}
\toprule
\multirow{2}{*}{Model} & \multicolumn{3}{c|}{\texttt{Trade}} & \multicolumn{3}{c}{\texttt{Genre}} \\
 & \name & DyGLib & TGB & \name & DyGLib & TGB \\
\midrule
P.F.                & \cfirst{0.06} & 1.35 & \csecond{0.15} & \cfirst{6.02} & 8.56 & \csecond{6.66} \\
TGN                 & \csecond{2.44} & 2.54 & \cfirst{2.19} & \cfirst{25.37} & 106.34 & \csecond{58.13} \\
DyGFormer           & \cfirst{3.49} & \csecond{21.13} & \multicolumn{1}{c|}{\cna} & \cfirst{11.78} & \csecond{588.69} & \multicolumn{1}{c}{\cna} \\
TGCN                & \cfirst{0.08} & \multicolumn{1}{c}{\cna} & \multicolumn{1}{c|}{\cna} & \cfirst{6.39} & \multicolumn{1}{c}{\cna} & \multicolumn{1}{c}{\cna} \\
GCLSTM              & \cfirst{0.07} & \multicolumn{1}{c}{\cna} & \multicolumn{1}{c|}{\cna} & \cfirst{6.48} & \multicolumn{1}{c}{\cna} & \multicolumn{1}{c}{\cna} \\
GCN                 & \cfirst{0.07} & \multicolumn{1}{c}{\cna} & \multicolumn{1}{c|}{\cna} & \cfirst{6.46} & \multicolumn{1}{c}{\cna} & \multicolumn{1}{c}{\cna} \\
\bottomrule
\end{tabular}
}
\label{tab:eval}
\end{table*}

\subsection{Peak GPU Usage}
\label{sec:gpu}
\noindent
\begin{minipage}{0.45\textwidth}  
Table~\ref{tab:gpu_usage_per_model} shows the peak GPU memory usage of each model across three standard datasets. Lightweight models such as GCN and GCLSTM consume minimal memory, making them efficient choices for resource-constrained environments, whereas larger architectures like GraphMixer and DyGFormer require significantly more GPU memory. This comparison highlights the trade-offs between model size and memory efficiency, providing a practical reference for selecting models in temporal graph learning tasks.
\end{minipage}%
\hfill
\begin{minipage}{0.50\textwidth}  
\centering
\captionof{table}{Peak GPU memory usage (GB) per model on different datasets.}
\resizebox{\linewidth}{!}{%
\begin{tabular}{lrrr}
\toprule
\textbf{Model} & \texttt{Wikipedia} & \texttt{Reddit} & \texttt{LastFM} \\
\midrule
TGAT       & 0.55 & 0.57 & 0.30 \\
TGN        & 0.67 & 0.81 & 0.11 \\
GraphMixer & 2.61 & 2.62 & 2.62 \\
DyGFormer  & 1.34 & 1.36 & 1.03 \\
TPNet      & 1.37 & 1.47 & 1.15 \\
GCLSTM     & 0.01 & 0.18 & 0.07 \\
GCN        & 0.01 & 0.09 & 0.05 \\
\bottomrule
\end{tabular}
}
\label{tab:gpu_usage_per_model}
\end{minipage}

\subsection{CProfiler Model Breakdown}
\label{sec:profiler}

Table \ref{tab:tgat_runtime_breakdown} shows a runtime decomposition of TGAT on the \texttt{LastFM} dataset. The largest costs arise from the backward pass (25.8\%), model forward (26.5\%), and optimizer updates (19.1\%), together accounting for over 70\% of total runtime. Within data loading (26.5\%), hook execution (15.1\%) and graph materialization (11.4\%) dominate, with the recency sampler alone contributing 13.2\%. Inside TGAT forward, attention layers (14.7\%) and MLPs (6.0\%) form the bulk of computation, while time encoding adds 3.5\%. Using a profiler in this way helps researchers and practitioners identify which components are the main bottlenecks and prioritize optimizations accordingly.

\definecolor{rowcolor}{HTML}{DDEBF7}

\begin{table}[ht]
\centering
\footnotesize
\caption{Breakdown of TGAT runtime on \texttt{LastFM} dataset.}
\label{tab:tgat_runtime_breakdown}
\begin{tabular}{lp{6cm}r}
\toprule
\textbf{Category} & \textbf{Component} & \textbf{Percent (\%)} \\
\midrule
\multirow{6}{*}{Data Loading} 
  & \cellcolor{rowcolor} Hook execution & \cellcolor{rowcolor}15.09 \\
  & \texttt{|--} Recency sampler & 13.19 \\
  & \texttt{|   |--} Get neighbors & 7.76 \\
  & \texttt{|   |--} Update circular buffer & 5.43 \\
  & \texttt{|--} Other hooks & 1.90 \\
  & \cellcolor{rowcolor} Graph materialization & \cellcolor{rowcolor} 11.40 \\
\midrule
\multirow{4}{*}{Model Forward} 
  & \cellcolor{rowcolor} TGAT forward & \cellcolor{rowcolor} 24.20 \\
  & \texttt{|--} Attention layers & 14.70 \\
  & \texttt{|--} MLP layers & 5.96 \\
  & \texttt{|--} Time encoding & 3.54 \\
  & \cellcolor{rowcolor} Other forward (decoders) & \cellcolor{rowcolor} 2.30 \\
\midrule
\multirow{4}{*}{Optimization} 
    & \cellcolor{rowcolor}Backward pass  & \cellcolor{rowcolor}25.80 \\
    & \cellcolor{rowcolor}Optimizer (Adam)  & \cellcolor{rowcolor} 19.10 \\
    & \cellcolor{rowcolor}Loss computation  & \cellcolor{rowcolor} 0.62 \\
\midrule
Other & - & 1.61 \\
\bottomrule
\end{tabular}
\end{table}

\subsection{Correctness Tests} \label{app:correctness}

Table \ref{tab:compact_performance_split} reports validation and MRR performance on \texttt{Wikipedia} for dynamic link property prediction, as well as validation NDCG and test NDCG on \texttt{Trade} for node property prediction. We cross-reference these results with TGB-reported performance and find that all models fall within the expected range. Note that we did not perform hyperparameter optimization, early stopping, or tuning, but instead used the hyperparameters listed in Table \ref{tab:hyperparams}.

\label{tab:compact_performance_split}

\begin{table}[ht]
\centering
\caption{Performance on \texttt{Wikipedia} and \texttt{Trade} datasets. Numbers are mean $\pm$ std over 3 runs.}
\label{tab:compact_performance_split}
\resizebox{\linewidth}{!}{
\begin{tabular}{llrrrr}
\toprule
\textbf{Category} & \textbf{Model} 
& \multicolumn{2}{c}{\texttt{Wikipedia}} 
& \multicolumn{2}{c}{\texttt{Trade}} \\
\cmidrule(lr){3-4} \cmidrule(lr){5-6}
 & & Validation MRR (↑) & Test MRR (↑) & Validation NDCG (↑) & Test NDCG (↑) \\
\midrule
\multirow{2}{*}{\textbf{Baselines}} 
 & Edgebank           & $0.495$ & $0.527$ & --- & --- \\
 & P.F. & --- & --- & $0.860$ & $0.855$ \\
\midrule
\multirow{3}{*}{\textbf{DTDG}} 
 & GCN       & $0.465 \pm$  \scriptsize{0.013} & $0.410 \pm$ \scriptsize{0.019} & \csecond{0.670 $\pm$ \scriptsize{0.013}} & \csecond{0.629 $\pm$ \scriptsize{0.009}} \\
 & GCLSTM    & $0.402 \pm$ \scriptsize{0.016} & $0.364 \pm$ \scriptsize{0.015} & \cfirst{0.761 $\pm$ \scriptsize{0.003}} & \cfirst{0.692 $\pm$ \scriptsize{0.002}} \\
 & TGCN      & --- & --- & $0.515 \pm $\scriptsize{0.006} & $0.458 \pm$ \scriptsize{0.007} \\
\midrule
\multirow{5}{*}{\textbf{CTDG}}
 & TGAT       & $0.380 \pm$ \scriptsize{0.013} & $0.322 \pm$ \scriptsize{0.013} & --- & --- \\
 & TGN        & $0.210 \pm$ \scriptsize{0.186} & $0.244 \pm$ \scriptsize{0.061} & $0.394$ & $0.329$ \\
 & GraphMixer & $0.610 \pm$ \scriptsize{0.010} & $0.567 \pm$ \scriptsize{0.018} & --- & --- \\
 & DyGFormer  & \csecond{0.743 $\pm$ \scriptsize{0.006}} & \csecond{0.712 $\pm$ \scriptsize{0.009}} & $0.386 \pm$ \scriptsize{0.0012} & $0.312 \pm$ \scriptsize{0.0003} \\
 & TPNet      & \cfirst{0.771 $\pm$ \scriptsize{0.033}} & \cfirst{0.747 $\pm$ \scriptsize{0.037}} & --- & --- \\
\bottomrule
\end{tabular}
}
\end{table}

\section{Additional Background: DTDG vs. CTDG}\label{app:DTDGvsCTDG}

As defined in Section~\ref{sec:tgm_frame}, a temporal graph is a graph whose structure and attributes evolve over time, capturing not only the relationships between entities but also the dynamics of their interactions. Unlike static graphs, which provide a single snapshot of connectivity, temporal graphs represent edges (and sometimes nodes) as time-stamped events or intervals, enabling modelling of when and how relationships form, change, or disappear. Temporal graph neural networks are typically categorized into two types: continuous-time dynamic graph (CTDG) methods and discrete-time dynamic graph (DTDG) methods. Section~\ref{app:DTDG} and Section~\ref{app:CTDG} provide further information about common approaches from each category.




\subsection{DTDG methods} \label{app:DTDG}

DTDG, or snapshot-based methods, take as input a sequence of graph snapshots, each representing the state of the temporal graph at discrete time intervals (e.g., hours or days). These approaches process each snapshot as a whole, typically using a graph learning model, and employ mechanisms to capture temporal dependencies across snapshots.

The majority of DTDG methods consist of two main components: a spatial encoder, commonly GNN-based, and a temporal encoder, usually an RNN or one of its variants. Given a snapshot $G_i$, a spatial representation is learned, $Z_i = f(V_i, E_i)$, where $f$ is a trainable or non-trainable function that takes the graph structure of the current snapshot and returns either node-level representations in $G_i$ or a representation of the entire snapshot. GCN~\citep{gcn} is used as $f$ in TGCN~\citep{t-gcn}, EvolveGCN~\citep{DBLP:conf/aaai/ParejaDCMSKKSL20}, and GCLSTM~\citep{gclstm}. In contrast, GraphPulse~\citep{DBLP:conf/iclr/ShamsiPHNCA24} encodes a whole-graph representation by extracting topological features from both the original graph $G_i$ and a transformed version $G_i'$, using Topological Data Analysis (TDA). The concatenation of the features from $G_i$ and $G_i'$ serves as the graph-level representation for downstream property prediction tasks.

To capture temporal dependencies across snapshots, an RNN or one of its variants (e.g., GRU or LSTM) is typically employed. These are applied either to the sequence of snapshot representations $Z_i$~\citep{t-gcn,gclstm,DBLP:conf/iclr/ShamsiPHNCA24} or directly to the evolving parameters of the GCN~\citep{DBLP:conf/aaai/ParejaDCMSKKSL20}.

\subsection{CTDG methods} \label{app:CTDG}

In contrast, CTDG methods operate on a continuous stream of edges and can make predictions at arbitrary timestamps. They update internal representations incrementally as new interactions arrive, incorporating fresh information into predictions. For computational efficiency, the edge stream is usually partitioned into fixed-size batches, with predictions performed sequentially per batch; once predictions are made, the corresponding edges are revealed to the model. Unlike DTDG methods, CTDG approaches do not rely on snapshots; instead, they maintain evolving node representations and sample temporal neighborhoods around nodes of interest for prediction.

\textbf{Temporal Message Passing.}
The temporal message passing framework is a neighbourhood aggregation scheme which recursively computes a latent representation by forwarding messages to temporal neighbours. Formally, if $\mathcal{N}^k(s)$ denotes the k-hop neighbourhood of node $s$ in the dynamic graph $\mathcal{G}$, then the \textit{temporal neighbourhood} $\mathcal{N}^k_t(s)$ is given by restricting neighbours to edge events chronologically before time $t$:
\begin{align}
    \mathcal{N}^k_t(s) = \{(s, d, t') \in \mathcal{N}^k(s) : t' \leq t \}
\end{align}

The combination of temporal and topological constraints makes efficient neighbourhood particularly challenging, requiring complex hierarchical data structures and cache-aware programming to sustain high-throughput on GPU stream multiprocessors \cite{lpma, gpma}. We bypass the insertion and deletion complexity by assuming the entire graph structure is read-only. Temporal message proceeds by creating and passing messages between such sub-neighorhoods. In particular, messages are created by concatenating embeddings, aggregating embeddings across temporal neighbourhoods, then updating the new hidden representation. Such information flow occurs concurrently for each event in a batch of data. 

\textbf{Time-Encoding and Memory-Based Learning.} Time-encoding models use a shift-invariant model $\psi: T \to \mathbb{R}^{d_t}$ that maps a real-valued time stamp into a $d_t$-dimensional vector (e.g. TGAT \cite{tgat} use time-encoders like Time2Vec \cite{time2vec}). This encoding is then passed through modified self-attention blocks or feedforward layers. \textit{Memory-based} models, such as TGN \cite{tgn}, utilize a fixed-bandwidth memory module that compresses relevant information for each node and updates it over time. EdgeBank \cite{edgebank} is a non-parametric, memory-based method that memorizes and predicts new links at test time based on their occurrence in the training data.


\section{Dataset Details} \label{app:data}

In this work, we conduct experiments on Wikipedia (obtained from the TGB~\cite{huang2023temporal}, where the dataset can be downloaded along with the package from \href{https://tgb.complexdatalab.com/}{TGB website}), Reddit, LastFM, datasets, obtained from~\cite{poursafaei2022towards}; these can be downloaded from \url{https://zenodo.org/records/7213796#.Y8QicOzMJB2}. 
These datasets span a variety of real-world domains, providing a broad testbed for evaluating temporal graph models.
Detailed information about these datasets are as follows.

\begin{itemize}[leftmargin=*, noitemsep, topsep=0.2pt]
    \item \textbf{Wikipedia} is a bipartite interaction network that captures editing activity on Wikipedia over one month. The nodes represent Wikipedia pages and their editors, and the edges indicate timestamped edits. Each edge is associated with a 172-dimensional LIWC feature vector derived from the text.  
    \item \textbf{Reddit} models user-subreddit posting behavior over one month. Nodes are users and subreddits, and edges represent posting requests made by users to subreddits, each associated with a timestamp. Each edge is associated with a 172-dimensional LIWC feature vector based on post contents.  
    \item \textbf{LastFM} is a bipartite user–item interaction graph where nodes represent users and songs. Edges indicate that a user listened to a particular song at a given time. The dataset includes 1000 users and the 1000 most-listened songs over a one-month period. 
    This dataset is not attributed.
    \item \textbf{Trade} represents the international agriculture trading network between UN nations from 1986 to 2016. Nodes are countries and edges capture the annual sum of agriculture trade values from one country to another. The task is to predict each nation’s trade proportions in the following year.
    \item \textbf{Genre} is a bipartite, weighted network connecting users to music genres based on listening history. Edges indicate the proportion of a song belonging to a genre that a user listens to, aggregated weekly. The task is to predict user-genre interactions in the next week, capturing evolving user preferences for music recommendation.
\end{itemize}

\begin{table*}[t]
\caption{Dataset statistics. } \label{tab:data} 
\centering
  \resizebox{0.9\linewidth}{!}{%
  \begin{tabular}{ l | rrrrrr}
  \toprule
  Dataset & \# Nodes & \# Edges
  & \# Unique Edges & \# Unique Steps & Surprise & Duration\\ 
  \midrule
  \texttt{Wikipedia} &  9,227 & 157,474 & 18,257 &  152,757 & 0.108 & 1 month\\ 
  \texttt{Reddit} & 10,984 & 672,447 & 78,516 & 669,065 & 0.069 & 1 month  \\  
  \texttt{LastFM} & 1,980 & 1,293,103 & 154,993 & 1,283,614 &  0.35 & 1 month \\
  \texttt{Trade} & 255 & 468,245 & 468,245 & 32 & 0.023 & 30 years \\
  \texttt{Genre} & 1,505 & 17,858,395 & 17,858,395 & 133,758 & 0.005 & 1 month \\

  \bottomrule
  \end{tabular}
  }
  \vspace{-10pt}
\end{table*}

\section{Temporal Graph Models Supported in \name}
\name is a research-driven library providing implementations of state-of-the-art temporal graph learning models. At the time of writing, TGM includes the following models:

\textbf{Persistent Forecast.} A simple baseline that predicts the future state of each node or edge by assuming it remains unchanged from the most recent observation. Despite its simplicity, it often serves as a strong baseline for dynamic node property prediction.

\textbf{EdgeBank.} \cite{edgebank} Maintains a memory bank of historical edges and uses them to make predictions. By storing and sampling past interactions, EdgeBank leverages temporal patterns without explicit node embedding updates, providing a lightweight but effective approach for dynamic link prediction.

\textbf{TGAT.} \cite{tgat} proposed to model dynamics node representations with TGAT layer, which is a combination of the graph attention mechanism with a time encoding function based on Bochner’s theorem, which provides a continuous functional mapping from time to a vector space. This allows TGAT to efficiently learn from temporal neighbourhood features with the aid of a self-attention mechanism and temporal dependencies encoded by the time encoding function.

\textbf{TGN.} \cite{tgn} proposed an event-based model that is a combination of a memory module, message aggregator, message updater and embedding module. In particular, the memory module maintains evolving memory for each node and updates this memory when the node is observed to be involved in an interaction, which is achieved by a message function, message aggregator, and message updater. Finally, the embedding model is used to compute the representation of nodes.

\textbf{GCN.}\citep{gcn} Standard Graph Convolutional Network applied on static snapshots to encode structural information. Node features are aggregated from neighbors and combined with self-features to produce updated embeddings at each snapshot. When used in temporal settings, GCNs process sequences of snapshots independently or in combination with temporal modules.

\textbf{GCLSTM.} To learn over a sequence of graph snapshots,~\cite{gclstm} proposed an end-to-end model named Graph Convolutional Long Short-Term Memory (GCLSTM) for dynamic link prediction. The LSTM serves as the backbone to capture temporal dependencies across graph snapshots, while a GCN is applied to each snapshot to encode structural dependencies between nodes. Specifically, two GCNs are used to update the hidden state and the cell state of the LSTM, and a multilayer perceptron (MLP) decoder maps the features at the current time step back to the graph space. This design enables GCLSTM to effectively handle both link additions and deletions.

\textbf{T-GCN.} \cite{t-gcn} integrates GCNs with gated recurrent units to learn node embeddings over sequences of graph snapshots, capturing temporal and structural information jointly.

\textbf{GraphMixer.} \citep{graphmixer} A graph adaptation of MLP-Mixer architectures. It alternates between node-wise and feature-wise mixing layers to capture structural correlations across nodes and temporal correlations across features. By stacking multiple mixer layers, GraphMixer can model complex dependencies in dynamic graphs while remaining simple and parameter-efficient.

\textbf{DyGFormer.} \cite{dygfomer} proposed a Transformer-based architecture for modeling dynamic graphs. DyGFormer consists of two key components: the Neighbour Co-occurrence Encoder and a Transformer. The Neighbour Co-occurrence Encoder leverages the recent first-hop neighbours of the source and destination nodes of an edge to capture correlations and compute relative embeddings. To enhance representation learning, \cite{dygfomer} further introduced a patching technique that splits the source and destination node features, edge features, time embeddings (computed following TGAT~\citep{tgat}), and relative embeddings into multiple patches. These patches are then fed into the Transformer to generate node representations with respect to an edge.

\textbf{TPNet.} TPNet is composed of two main modules: Node Representation Maintenance and Link Likelihood Computation. \cite{tpnet} unifies existing relative encoding methods by introducing temporal walk matrices with an integrated time-decay function. These matrices establish a principled connection between relative encodings and temporal walks, offering a clearer framework for analyzing and designing temporal encodings. The time-decay effect further allows joint modelling of temporal and structural information. Since computing temporal walk matrices directly is computationally and memory intensive, TPNet employs a theoretically grounded random feature propagation mechanism to implicitly approximate and maintain them efficiently.

The \name team is actively expanding the library to incorporate additional cutting-edge models, including TNCN~\citep{tncn}, DyGMamba~\citep{dygmamba}, NAT~\citep{nat}, and TGNv2~\citep{tgnv2}.

\section{Compute Resources and Experiment Details}
\label{app:details}

\textbf{Compute}: Experiments were run on Ubuntu 20.04 with 64 GB RAM, 4 isolated AMD EPYC 7502 CPU cores, and a single 80 GB A100 GPU. Jobs were managed with SLURM to ensure isolated environments and no concurrent interference.

\textbf{Experiment Details}: We use the default TGB splits~\citep{huang2023temporal, tgb2}, with hyperparameters listed in Table~\ref{tab:hyperparams}. For efficiency benchmarks, TGAT and TGN adopt the TGLite configuration~\citep{10.1145/3620665.3640414} for fairness. Other libraries were modified only minimally to measure latency, and TGLite/TGL times are taken directly from Fig. 6 of~\citep{10.1145/3620665.3640414}. All DTDG methods discretized the \texttt{Trade} dataset to yearly snapshots, and the \texttt{Genre} dataset to weekly snapshots.

\begin{table*}[h!]
\centering
\scriptsize
\setlength{\tabcolsep}{4pt}
\caption{Hyperparameters used for each model}
\begin{tabular}{lrrrrrrrrr}
\toprule
\textbf{Parameter} & \textbf{Edgebank} & \textbf{TGAT} & \textbf{TGN} & \textbf{GCN} & \textbf{GCLSTM} & \textbf{TGCN} & \textbf{GraphMixer} & \textbf{DyGFormer} & \textbf{TPNet} \\
\midrule
Batch Size             & 200 & 200 & 200 & 200 & 200 & – & 200 & 200 & 200 \\
Epochs                 & –   & 10  & 30  & 30  & 30  & – & 10  & 5   & 10 \\
Learning Rate          & –   & 1e-4& 1e-4& 1e-3& 1e-3& 1e-3 & 2e-4& 1e-4& 1e-4 \\
Dropout                & –   & 0.1 & 0.1 & 0.1 & –   & 0.1 & 0.1 & 0.1 & 0.1 \\
\# Heads               & –   & 2   & 2   & –   & –   & –   & –   & 2   & – \\
\# Neighbors           & –   & 20  & 10  & –   & –   & –   & 20  & 32  & 32 \\
\# Layers              & –   & 2   & 2   & 2   & 2   & 2   & 2   & 2   & 2 \\
Embedding Dim.         & –   & 100 & 100 & 128 & 256 & 128 & 128 & 172 & 172 \\
Time Dim.              & –   & 100 & 100 & –   & –   & –   & 100 & 100 & 100 \\
Memory Dim.            & –   & –   & 100 & –   & –   & –   & –   & –   & – \\
Node Dim.              & –   & –   & –   & 256 & 256 & 256 & 100 & 128 & 128 \\
Sampling               & –   & Recency & Recency & – & – & – & Recency & Recency & Recency \\
Memory Mode            & Unlimited & – & – & – & – & – & – & – & – \\
Time Gap               & –   & –   & –   & –   & –   & –   & 2000 & – & – \\
Token Dim. Factor      & –   & –   & –   & –   & –   & –   & 0.5 & – & – \\
Channel Dim. Factor    & –   & –   & –   & –   & –   & –   & 4.0 & – & – \\
Channel Dim.           & –   & –   & –   & –   & –   & –   & –   & 50  & – \\
Patch Size             & –   & –   & –   & –   & –   & –   & –   & 1   & – \\
\# Channels            & –   & –   & –   & –   & –   & –   & –   & 4   & – \\
\# RP Layers           & –   & –   & –   & –   & –   & –   & –   & –   & 2 \\
RP Time Decay          & –   & –   & –   & –   & –   & –   & –   & –   & 1e-6 \\
RP Dim                 & –   & –   & –   & –   & –   & –   & –   & –   & $\log(|2 * E|)$ \\
\bottomrule
\end{tabular}
\label{tab:hyperparams}
\end{table*}

\end{document}